\pdfoutput=1
\documentclass[11pt]{article}

\usepackage{acl}

\usepackage{times}
\usepackage{latexsym}

\usepackage[T1]{fontenc}
\usepackage[utf8]{inputenc}

\usepackage{microtype}

\usepackage{booktabs}
\usepackage{multirow}
\usepackage{graphicx}
\usepackage{xspace}
\usepackage{comment}
\usepackage{amsmath}
\usepackage{amssymb}
\usepackage{verbatim}
\usepackage{fancyvrb}
\usepackage{verbatimbox}
\usepackage{xcolor}

\definecolor{lightmauve}{rgb}{0.86, 0.82, 1.0}
\definecolor{lightsalmon}{rgb}{1.0, 0.63, 0.48}

\newcommand{\bart}[0]{BART\xspace}%{\includegraphics[scale=0.08]{pics/bart.png}\xspace}
\newcommand{\sema}[0]{\textsc{Sema}\xspace}
\newcommand{\sembleu}[0]{\textsc{SemBleu}\xspace}
\newcommand{\smatch}[0]{\textsc{Smatch}\xspace}
\newcommand{\sxmatch}[0]{\textsc{S$^2$match}\xspace}
\newcommand{\wlk}[0]{\textsc{Wlk}\xspace}
\newcommand{\wwlk}[0]{\textsc{Wwlk}\xspace}
\newcommand{\simple}[0]{\textsc{Simple}\xspace}

\title{Better Smatch = Better Parser? AMR evaluation is not so simple anymore}

\author{Juri Opitz \\
  Dept.\ of Computational Linguistics \\
  Heidelberg University \\
  69120 Heidelberg \\
  \texttt{opitz.sci@gmail.com} \\\And
  Anette Frank \\
  Dept.\ of Computational Linguistics \\
  Heidelberg University \\
  69120 Heidelberg \\
  \texttt{frank@cl.uni-heidelberg.de} \\}

\begin{document}
\maketitle
\begin{abstract}
Recently, astonishing advances have been observed in AMR parsing, as measured by the structural \smatch metric. In fact, today's systems achieve performance levels that seem to surpass estimates of human inter annotator agreement (IAA). Therefore, it is unclear how well \smatch (still) relates to human estimates of parse quality, as in this situation potentially fine-grained errors of similar weight may impact the AMR's meaning to different degrees. 

We conduct an analysis of two popular and strong AMR parsers that -- according to \smatch -- reach quality levels on par with  human IAA, and assess how human quality ratings relate to \smatch and other AMR metrics. Our main findings are: i) While high \smatch scores indicate otherwise, we find that \textbf{AMR parsing is far from being solved}: we frequently find structurally small, but semantically unacceptable errors that substantially distort sentence meaning. ii) Considering high-performance parsers, \textbf{better \smatch scores may not necessarily indicate consistently better parsing quality}. To obtain a meaningful and comprehensive assessment of quality differences of parse(r)s, we recommend augmenting evaluations with macro statistics, use of additional metrics, and more human analysis. 
\end{abstract}

\section{Introduction}

Abstract Meaning Representation (AMR), proposed by \citet{banarescu-etal-2013-abstract}, aims at capturing the meaning of texts in an explicit graph format. Nodes describe \textit{entities, events}, and \textit{states}, while edges express key semantic relations, such as \textit{ARG}$_x$ (indicating semantic roles as in PropBank \cite{palmer-etal-2005-proposition}), or \textit{instrument} and \textit{cause}.

\begin{figure}[t]
    \centering
    \includegraphics[width=0.8\linewidth, trim={0 1cm 0 0},clip]{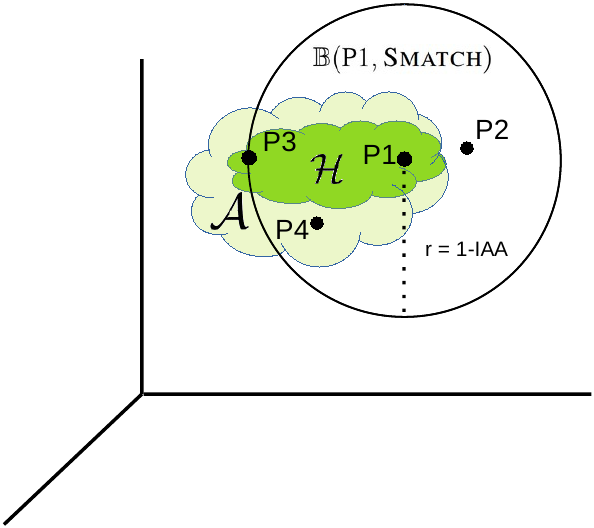}
    \caption{Sketch of AMR IAA ball. The center (P1) is a reference AMR, while P2, P3, P4 are candidates. Any AMR $x$ from the ball has high structural \smatch agreement with P1, i.e., $\smatch(x,P1)$ $\geq$ estimated human IAA. However, they may fall in different categories: $\mathcal{H}$ (green cloud) contains correct AMR alternatives. Its superset $\mathcal{A}$ (light cloud) contains acceptable AMRs that may misrepresent 
    the sentence meaning up to a minor degree. Other parses from the ball, e.g., P2, mis-represent the sentence's meaning -- despite possibly having higher \smatch agreement with the reference than all other candidates.
    }
    \label{fig:motivate}
\end{figure}

Albeit the development of parsers can be driven by multiple desiderata, better performance on benchmarks often serves as main criterion. For AMR, this goal is typically measured using \smatch \cite{cai-knight-2013-smatch} against a reference corpus. The metric measures to what extent the reference has been reconstructed by the parser.

However, thanks to astonishing recent advances in AMR parsing, mainly powered by the \textit{language modeling and fine-tuning paradigm} \cite{bevilacqua2021one}, parsers now achieve benchmark scores that surpass IAA estimates.\footnote{\citet{banarescu-etal-2013-abstract} find that an (optimistic) average annotator vs.\ consensus IAA (\smatch) was 0.83 for newswire and 0.79 for web text. When newly trained annotators doubly annotated web text sentences, their annotator vs.\ annotator IAA was 0.71. Recent BART and T5 based models range between 0.82 and 0.84 \smatch F1 scores.} Therefore, it is difficult to assess whether (fine) differences in \smatch scores i) can be attributed to minor but valid divergences in interpretation or  AMR structure, as they may also occur in human assessments, or ii) if they constitute significant meaning distorting errors. 

This fundamental issue is outlined in Figure \ref{fig:motivate}. Four parses are located in the ball $\mathbb{B}(\text{P1}, \smatch)$ of estimated IAA, (gold) parse P1 being the center. However, the true set of possible human candidates $\mathcal{H}$ is very likely much smaller than the ball and its shape is unknown.\footnote{Under the unrealistic assumptions of an omniscient annotator and AMR being the ideal way of meaning representation, one might require that $\mathcal{H}$ always has exactly one element.} Besides, a superset of $\mathcal{H}$ is a set of \textit{acceptable} parses $\mathcal{A}$, i.e., parses that may have a small flaw which does not significantly distort the sentence meaning. Now, it can indeed happen that parse P2, as opposed to P3, has a lower  distance to reference P1, i.e., to the center of $\mathbb{B}(\smatch)$ -- but is not found in $\mathcal{A} \supseteq \mathcal{H}$, which marks it as an inaccurate candidate. On the other hand, P4 is contained in $\mathcal{A}$, but  not in $\mathcal{H}$, which would make it acceptable, but less preferable than P3. 

\paragraph{Research questions} Triggered by these considerations, this paper tackles the key questions: \textit{Do high-performance AMR parsers indeed deliver accurate semantic graphs, as suggested by high benchmark scores that surpass human IAA estimates}? \textit{Does a higher \smatch against a single reference necessarily indicate better overall parse quality}? 
And \textit{what steps can we take to mitigate potential issues when assessing the true performance of high-performance parsers?} 

\paragraph{Paper structure} After discussing background and related work (Section \ref{sec:background}), we describe our data setup and give a survey of AMR metrics (Section \ref{sec:setup}). We then evaluate the metrics with regard to scoring i) corpora (Section \ref{sec:syslevelscoring}), ii) AMR pairs (Section \ref{sec:metricaccuracy}) and iii) cross-metric differences in their ranking behavior (Section \ref{sec:metricspecificity}). We conclude by discussing limitations of our study (Section \ref{sec:limitations}), give recommendations and outline future work (Section \ref{sec:recommendations}).\footnote{Code and data for our study are available at \url{https://github.com/Heidelberg-nlp/AMRParseEval}.}

\section{Background and related work}
\label{sec:background}

\paragraph{AMR parsing and applications} Over the years, we have observed a great diversity in approaches to AMR parsing, ranging from graph prediction with a pipeline \cite{flanigan-etal-2014-discriminative}, or a neural network \cite{lyu-titov-2018-amr, cai-lam-2020-amr} to transition-based parsing \cite{wang-etal-2015-transition} and sequence-to-sequence parsing, e.g., by exploiting large parallel corpora \cite{xu-etal-2020-improving}. A recent trend is to exploit the knowledge in large pre-trained sequence-to-sequence language models such as T5 \cite{T5} or BART \cite{lewis-etal-2020-bart}, by fine-tuning them on AMR corpora, as show-cased, e.g., by  \citet{bevilacqua2021one}. Such models are on par or tend to surpass estimates for human AMR agreement \cite{banarescu-etal-2013-abstract}, when measured in \smatch points.

AMR, by virtue of its properties as a graph-based abstract meaning representation, is attractive for many key NLP tasks, such as machine translation \cite{10.1162/tacl_a_00252}, summarization \cite{dohare2017text, liao-etal-2018-abstract}, NLG evaluation \cite{opitz2020towards, manning-schneider-2021-referenceless, ribeiro2022factgraph} and measuring semantic sentence similarity \cite{opitz2022sbert}. 

\paragraph{Metric evaluation for MT evaluation} Metric evaluation for machine translation (MT) has received much attention over the recent years \cite{ma-etal-2019-results, mathur-etal-2020-results, freitag-etal-2021-results}. When evaluating metrics for MT evaluation, it seems generally agreed upon that the main goal of a MT metric is high correlation to human ratings, mainly with respect to rating adequacy of a candidate against one (or a set of) gold reference(s). 

A recent shared task \cite{freitag-etal-2021-results} meta-evaluates popular metrics such as BLEU \cite{papineni-etal-2002-bleu} or BLEURT \cite{sellam-etal-2020-bleurt}, by comparing the metrics' scores to human scores for systems and individual segments. They find that the performance of each metric varies depending on the underlying domain (e.g., TED talks or news), and that most metrics struggle to penalize translations with errors in reversing negation or sentiment polarity, and show lower correlations for semantic phenomena including subordination, named entities and terminology. This indicates that there is potential for cross-pollination: clearly, AMR metric evaluation may profit from the vast amount of experience of metric evaluation for other tasks. On the other hand, MT evaluation may profit from relating semantic representations, to better differentiate semantic errors with respect to their type and severity. A first step in this direction may have been made by \citet{zeidler-etal-2022-dynamic}, who assess the behaviour of MT metrics, AMR metrics, and hybrid metrics when analyzing sentence pairs that differ in only one linguistic phenomenon.

\section{Study Setup: Data creation and AMR metric overview}
\label{sec:setup}

In this Section, first we select data and two popular high-performance parsers for creating candidate AMRs. Then we describe the human quality annotation, and give an overview of automatic AMR metrics that we consider in our subsequent studies.

\paragraph{Parsers and corpora}

We choose the AMR3 benchmark\footnote{LDC corpus  \url{LDC2020T02}} and the literary texts from the freely available Little Prince corpus.\footnote{From \url{https://amr.isi.edu/download.html}} As parsers we choose T5- and BART-based systems, both on par with human IAA estimates, where \bart achieves higher scores on AMR3.\footnote{See \url{https://github.com/bjascob/amrlib-models} for more benchmarking statistics.}  We proceed as follows: we 1.\ parse the corpora with T5 and BART parsers and use \smatch to select diverging parse candidate pairs, and 2.\ sample 200 of those pairs, both for AMR3, and for Little Prince (i.e., 800 AMR candidates in total).

\subsection{Annotation dimensions}

\paragraph{Annotation dimension I: pairwise ranking} The annotator is presented the sentence and two candidate graphs, assigning one of three labels and a free-text rationale. The labels are either +1 (prefer first graph), $-1$ (prefer second graph), or 0 (both are of same or very similar quality).

\paragraph{Annotation dimension II: parse acceptability} In addition, each graph is independently assigned a single label, considering only the sentence that it is supposed to represent. Here, the annotator makes a binary decision: +1, if the parse is acceptable, or 0, if the graph is not acceptable. A graph that is acceptable is fully valid, or may allow a very minor meaning deviation from the sentence, or a slightly weird but allowed interpretation that may differ from a normative interpretation. All other graphs are deemed not acceptable (0).

\paragraph{Example: Acceptable candidates, low \smatch }

\begin{figure}
\begin{tiny}
\begin{Verbatim}[commandchars=\\\{\},codes={\catcode`$=3\catcode`^=7}]
\textbf{-----------------Reference AMR and Sentence------------------}
(l / look-over-06               ``Looking over to the flag''
      :ARG1 (f / flag))
\textbf{---------------------Candidate parses------------------------}
(l / look-01                    (z0 / look-01
      :direction (o / over)           :ARG2 (z1 / flag)
      :destination (f / flag))        :direction (z2 / over))
\textbf{---------------------------Eval------------------------------}
  \colorbox{lightsalmon}{Smatch}(ref, cand): both score 0.2 (indicates low quality)
  \colorbox{lightmauve}{Human}(sent, cand): both are acceptable
  \colorbox{lightmauve}{Human}(cand, cand): no preference
\textbf{-------------------------------------------------------------}
   \end{Verbatim}
   \end{tiny}
     \vspace{-4mm}
\caption{Data example: acceptable, low \smatch. That is, $P \in \mathcal{H}$ but $P \notin \mathbb{B}(\smatch, ref)$.}
\label{fig:equalquality}
\end{figure}

\begin{figure}
\begin{tiny}
\begin{Verbatim}[commandchars=\\\{\},codes={\catcode`$=3\catcode`^=7}]
\textbf{----------------Reference AMR (excerpt)--------------------}
(i2 / imagine-01
     :ARG0 (y / you)
     :ARG1 (a / amaze-01
            :ARG1 (i / i)))
            :time-of (w / \textbf{wake-01}
                  :ARG0 (v / voice
                       :mod (o / odd)
                       :mod (l / little))
                  :ARG1 i))))))
\textbf{----------------Candidate parse (excerpt)--------------------}
(ii / imagine-01
     :ARG0 (y / you)
     :ARG1 (a / amaze-01
            :ARG0 (v / voice
                    :mod (l / little)
                    :mod (o / odd))
            :ARG1 (ii2 / i)))
                        
\textbf{Means:} (..) imagine my amazement (..) by an odd little voice
\textbf{Should mean:} (..) imagine my amazement (..) when I was 
             \textbf{awakened} by an odd little voice 
\textbf{---------------------------Eval------------------------------}
  \colorbox{lightsalmon}{Smatch}(ref, cand): scores 0.88 (indicates high quality)
  \colorbox{lightmauve}{Human}(sent, cand): not acceptable
\textbf{-------------------------------------------------------------}
   \end{Verbatim}
   \end{tiny}
  \vspace{-4mm}
\caption{Data example excerpt that shows an unaccaptable parse with high \smatch. That is, $P \not\in \mathcal{A} \supseteq \mathcal{H}$ but $P \in \mathbb{B}(\smatch, ref)$}
\label{fig:equalquality2}
\end{figure}

Figure \ref{fig:equalquality} shows an example of two graphs that have very low structural overlap with the reference (\smatch=~0.2), but are acceptable. Here, the candidate graphs both differ from the reference because they tend to a more conservative interpretation, using the more general \textit{look-01} predicate instead of the \textit{look-over-06} predicate in the human reference. In fact, the meaning of the reference can be considered, albeit valid, slightly weird, since \textit{look-over-06} is defined in PropBank as \textit{examining something idly}, which is a more `specific' interpretation of the sentence in question. On the other hand, the candidate graphs differ from each other in the semantic role assigned to \textit{flag}. In the first, \textit{flag} is the destination of the \textit{looking} action (which can be accepted), while in the second, we find a more questionable but still acceptable interpretation that \textit{flag} is an \textit{attribute of the thing that is looked at}. 

\paragraph{Example: Candidate not acceptable, high \smatch } An inverse example (high \smatch, unacceptable) is shown in Figure \ref{fig:equalquality2}, where the parse omits \textit{awaken}. Albeit the factuality of the sentence is not (much) changed, and the structural deviation may legitimately imply that the odd voice is the cause of amazement, it misses a relevant piece of meaning and is therefore rated unacceptable.

\paragraph{Label statistics} will be discussed in Section \ref{sec:syslevelscoring}, where the human annotations are also contrasted against parser rankings of automatic metrics.

\subsection{Metric overview}
\label{sec:overview}

We distinguish metrics targeting \textit{monolingual AMR parsing evaluation} from \textit{multi-purpose AMR metrics}. AMR metrics that are designed for evaluation of monolingual parsers typically have two features in common. First, they compare a candidate against a reference parse that both (try to) represent the \textit{same sentence}. Second, they measure the amount of successfully reconstructed reference structure.\footnote{The notion of \textit{success} is mostly focused on structural matches, and can vary among metrics, usually depending on theoretical arguments of the developers of the metric.}  

We also consider multi-purpose AMR metrics that aim to extend to use cases where AMRs represent \textit{different sentences}, such as evaluation of cross-lingual AMR parsing, natural language generation (NLG) or rating semantic sentence similarity.

\subsubsection{Monolingual AMR parsing metrics}

\paragraph{Triple matching strategies} \smatch \cite{cai-knight-2013-smatch} and \sema \cite{anchieta2019sema} consider graph triples as the elementary constituents of an AMR graph. Both compute a triple overlap score between candidate and reference parses. \smatch computes an  alignment between the variable nodes of two AMRs, which is accurate but slow. The \sema metric achieves a large speed-up by removing AMR variables from the graphs, replacing them with concept labels. 

\paragraph{Inspired by BLEU: \sembleu} BLEU \cite{papineni-etal-2002-bleu} is a popular (but debated) metric for machine translation evaluation. It matches bags-of-k-grams from candidate and reference, with a geometric mean of the precision scores over the $k$ different bags. Inspired by BLEU, and, similar to \sema, driven by the goal to make AMR evaluation more fast and efficient, \citet{song-gildea-2019-sembleu} propose the \sembleu metric for AMR graphs. It  extracts bags-of-k-grams from graphs, collected via breadth-first traversal. A point of motivation, similarly to \sema, is that the metric skips the costly graph alignment. Per default, \sembleu uses $k$=$3$. In this work we additionally use $k$=$2$, following \citet{opitz2021weisfeiler} who find that $k$=$2$ better relates to human notions of sentence similarity.

\subsubsection{Multi-purpose metrics}

\paragraph{\sxmatch and \wlk/\wwlk} Targeting AMR metric application cases beyond monolingual parsing evaluation, such as measuring AMR similarity of different sentences, or cross-lingual AMR parsing evaluation,  \citet{opitz-tacl, opitz2021weisfeiler, uhrig-etal-2021-translate} propose three metrics: i) \textbf{\sxmatch} is an adaption of \smatch that computes graded concept similarity (reflecting that, e.g., \textit{cat} is more similar to \textit{kitten} than to \textit{plant}). ii) \textbf{\wlk} applies the Weisfeiler-Leman kernel \cite{shervashidze2011weisfeiler} to compute a similarity score over feature vectors that describe graph statistics in different iterations of node contextualization. iii) \textbf{\wwlk} (Wasserstein \wlk, \citet{NEURIPS2019_73fed7fd}) projects the nodes of the graphs to a latent space partitioned into different degrees of node contextualization. Wasserstein distance is then used to match the graphs, based on a pair-wise node distance matrix.

\paragraph{Setup of multi-purpose metrics} For \sxmatch, \wlk and \wwlk we use the default setup, which consists of GloVe \cite{pennington-etal-2014-glove} embeddings and $k$=$2$ in \wlk and \wwlk, where $k$ indicates the depth of node contextualizations.

\textbf{Default \wwlk} initializes parameters randomly, if tokens are out of vocabulary (a random embedding for each OOV token type). To achieve deterministic results, without fixing a random seed, we could initialize the OOV parameters to 0. However, with this we'd lose valuable discriminative information on graph similarity. We therefore adopt a slight adaptation for \wwlk and calculate the \textit{expected} distance matrix before Wasserstein metric calculation, making results more reproducible while keeping discriminative power. 

We also introduce \wwlk-\textbf{k3e2n}, a \wwlk variant with \textit{edge2node (e2n)} transforms, more tailored to monolingual AMR parsing evaluation, which is the focus of this paper. It increases the score impact of edge labels, motivated by the insight that edge labels are of particular importance in AMR parsing evaluation. It transforms an edge-labeled graph into an \textit{equivalent} graph without edge-labels.\footnote{E.g., $(x, arg0, z) \rightarrow (x, y) \land (y, z) \land (y, arg0)$.} This is also known as `Levi transform' \cite{levi1942finite}, and has been previously advocated for AMR representation by \citet{beck-etal-2018-graph} and \citet{ribeiro-etal-2019-enhancing}. Since due to the transform the distances in the graph will grow, we increase $k$ by one (\textbf{k=3}). With this, we can set all edge weights to 1. 

\subsubsection{Simple baseline}

To put the results into perspective, we introduce a very \simple baseline: \simple extracts bag-of-words (relation and concept labels) from two AMR graphs and computes the size of their intersection vs.\ the size of their union (aka \textit{Jaccard Coefficient}).

\section{Preliminaries}
\label{sec:prelims}

We denote an AMR metric $m$ over AMRs  as:

\begin{equation}
    m: \mathcal{A} \times \mathcal{A} \rightarrow \mathbb{R},
\end{equation}

and a human metric $h$ as

\begin{equation}
    h: \mathcal{A} \times \mathcal{S} \rightarrow \mathbb{R},
\end{equation}

where $\mathcal{S}$ contains sentences. 

\section{Study I: System-level scoring}
\label{sec:syslevelscoring}

\paragraph{Research questions} We focus on two questions:

\begin{enumerate}
    \item How are the two parsers rated by humans? 
    \item How do metrics score our two parsers?
\end{enumerate}

With 1.\ we aim to assess whether there is still room for AMR parser improvement, even though their \smatch scores pass estimated human IAA. And for 2.\ we aim to know whether the metric rankings (still) appropriately reflect parser quality.

\subsection{System scoring}

\paragraph{Aggregation strategies: Micro vs.\ Macro} We have defined a metric between two AMRs. For ranking systems, we need to aggregate the individual pair-wise assessments into a single score. At this point, it is important to note that most papers use (only) micro \smatch for ranking parsers, i.e., counting triple matches of aligned AMR pairs over all AMR pairs (before a final F1 score calculation). 

Naturally, such micro corpus statistics are \textit{unbiased} w.r.t.\ to whatever is defined as a single evaluation instance (in \smatch: triples), but the trade-off is that they are biased towards instance type frequency and sentence length, since longer sentences tend to yield substantially more triples. Hence, the influence of a longer sentence may marginalize the influence of a shorter sentence. This issue may be further aggravated by the fact that longer sentences tend to contain more named entity phrases, and entity phrases typically trigger large simple structures, that are mostly easy to project.\footnote{As a small example, consider \textit{The bird sings} vs.\ \textit{Jon Bon Jovi sings}. The first sentence yields 3 triples, while the second sentence yields 8 triples, where the \textit{John Bon Jovi} named entitiy structure has added 6 triples, outweighing the key semantic event \textit{x sings}. Micro score would assign 2.6 times more importance to the second sentence/AMR.} Therefore, micro corpus statistics alone \textit{could} potentially yield an incomplete assessment of parser performance. To shed more light on this issue, we provide additional evaluation via macro aggregation.

\paragraph{Statistics for micro and macro system scoring} We calculate two statistics. The first statistic shows the (micro/macro)-aggregated corpus score for a metric $m$, parsed corpus $X$ and gold corpus $G$:
\begin{align*}
&\mathbb{S}(m, X, G) \\
&=AGGR(\{m(X_1, G_1), ... , m(X_n, G_n)\}),
\end{align*}

For macro metrics,  $AGGR$ is the mean of pair-wise scores over all instances in a corpus $X$. In case of the human metric, this is the ratio of acceptable parses in $X$. For micro metrics, $AGGR$ computes overall matching triple F1 (\smatch, \sema) or overall k-gram BLEU (\sembleu). For \wlk and \wwlk, a micro variant is not implemented, hence we only show their macro scores.

The second statistic shows how often $m$ prefers the parses in a parse corpus $X$  over the these in $Y$:

\begin{align*}
    \mathbb{P}(m, X, Y, G) = \sum_{i=1}^n \mathbb{I}[m(X_i, G_i) > m(Y_i, G_i)].
\end{align*}

Here, $\mathbb{I}[c]$ denotes a function that returns $1$ if the condition $c$ is true, and zero in all other cases. For better comparability of numbers, we distribute cases where $m(X_i, G_i) = m(Y_i, G_i)$, which are frequent for the human metric, evenly over $\mathbb{P}(m, X, Y, G)$ and $\mathbb{P}(m, Y, X, G)$.

\subsection{Results}

\begin{table*}
    \centering
    \scalebox{0.85}{
    \begin{tabular}{|ll|rrr|rrr|rrr|rrr|}
    \toprule
   & &\multicolumn{6}{c|}{Little Prince}& \multicolumn{6}{c|}{AMR3}\\
   & & \multicolumn{3}{c|}{$\mathbb{P}$}& \multicolumn{3}{c|}{$\mathbb{S}$}& \multicolumn{3}{c|}{$\mathbb{P}$}& \multicolumn{3}{c|}{$\mathbb{S}$} \\
   \cmidrule{3-14}
        & &  \bart & T5 &$\Delta$&   \bart & T5 &$\Delta$ & \bart &T5 & $\Delta$ & \bart &  T5 & $\Delta$\\
         \toprule
         \multirow{11}{*}{\rotatebox{90}{Macro}} 
       & HUM  & 87 & 113 & -26 & 0.58 & 0.69 & -0.11 & 100 & 100 & 0.0  & 0.62 & 0.62 & 0.00\\
       \cmidrule{2-14}
         &\simple & 87 & 113 & -26 & 0.69 & 0.7 & -0.01 & 82 & 118 & -36 & 0.75 & 0.75 & 0.00 \\
         \cmidrule{2-14}
       &  \sema & 84 & 116 & -32 & 0.6 & 0.63 & -0.03& 89 & 111 & -22 & 0.68 & 0.68 & 0.00\\
       &  \sembleu-k2 & 90 & 110 & -20 & 0.61 & 0.63 & -0.02 & 98 & 102 & -4 & 0.70 & 0.69 & 0.01 \\
       &  \sembleu-k3 & 90 & 110 & -20 & 0.51 & 0.53 & -0.02& 103 & 97 & 6 & 0.58 & 0.58 & 0.00\\
       &  \smatch &94 & 106 & -12 & 0.73 & 0.74 & -0.01& 95 & 105 & -10 & 0.77 & 0.77 & 0.00\\
       &  \sxmatch & 93 & 107 & -14 & 0.75 & 0.76 & -0.01&95 & 105 & -10 & 0.79 & 0.79 & 0.00\\
       &  \wlk-k2 & 92 & 108 & -16 & 0.63 & 0.65 & -0.02 & 96 & 104 & -8 & 0.69 & 0.69 & 0.00\\
       &  \wwlk-k2&91 & 109 & -18 & 0.79 & 0.8 & -0.01 & 102 & 98 & 4 & 0.84 & 0.84 & 0.00\\
       & \wwlk-k3e2n & 97 & 103 & -6 & 0.72 & 0.73 & -0.01 & 94 & 106 & -12 & 0.78 & 0.78 & 0.00 \\
         \midrule
         \multirow{4}{*}{\rotatebox{90}{Micro}} 
     & \sema &-&-&-&0.62&0.64& -0.02 &-&-&-& 0.69 & 0.68 & 0.01 \\
      & \sembleu &-&-&-&0.53&0.54& -0.01 &-&-&-& 0.60&0.57& 0.03  \\
      &\smatch &-&-&-&0.74&0.74& -0.01 &-&-&-& 0.77&0.75& 0.02 \\
      & \sxmatch &-&-&-&0.76&0.76& 0.00 &-&-&-& 0.80&0.77& 0.03 \\
\bottomrule
    \end{tabular}}
    \caption{Corpus level scoring results. Negative $\Delta$ shows preference for T5, positive $\Delta$ shows preference for BART.}
    \label{tab:parser_rank}
\end{table*}

Results are shown in Table \ref{tab:parser_rank}. In view of our research questions, we make interesting observations.

\paragraph{AMR parsing is far from solved} Considering the ratio of parses that were rated acceptable by the human (HUM, $\mathbb{S}$), they are surprisingly low, at only 0.58 (\bart, Little Prince, Table \ref{tab:parser_rank}); 0.69 (T5, Little Prince). Other parses have errors that substantially distort sentence meaning, even though major parts of the AMRs may structurally overlap.

\paragraph{Better \smatch on AMR benchmark may not (always) imply a better parser} On AMR3, when inspecting corpus-\smatch  (micro \smatch, Table \ref{tab:parser_rank}), \bart is considered the better parser, in comparison to T5 (+2 points). However, when consulting macro statistics, a different picture emerges. Here, \bart and T5 obtain the same scores: AMR3, 0.62 vs. 0.62, Table \ref{tab:parser_rank}. On the literary texts (Little Prince), where the domain is different and sentences tend to be shorter, T5 significantly (binomial test, $p<$0.05) outperforms \bart, both in the ratio of acceptable sentences (\bart: 0.58, T5: 0.69), and in number of preferred candidates (\bart: 87, T5: 113). Note that this insight is independent from our human annotations.

All in all, this may suggest that \bart tends to provide better performance for longer sentences, while T5 tends to provide better performance especially for shorter and medium-length sentences. Further analysis provides more evidence for this, cf.\ Appendix \ref{app:sentlen}: Figure \ref{fig:slenhum} and Figure \ref{fig:slensm}).

\paragraph{Metrics for system ranking} Regarding our tested metrics, especially the macro metrics, a clear pattern is that they mostly agree with the human ranking.  However, our current  results for the different metrics do not tell much, yet, about their suitability for AMR assessment and ranking. Even if a metric ranks a parser more similarly to the human, this may be for the wrong reasons, since this statistic filters out pair-wise correspondences to the human. This is also indicated by results of the simplistic bag-of-structure metric \simple, which achieves the same results as human (HUM) on Little Prince, with respect to the number of preferred parses ($\mathbb{P}$, Little Prince, Table \ref{tab:parser_rank}, HUM vs.\ \simple). In that respect, it is more important to assess the pair-wise metric accuracy and metric specificity, which we will visit next in Sections \ref{sec:metricaccuracy} and \ref{sec:metricspecificity}.

\section{Study II: Metric accuracy on parse level}
\label{sec:metricaccuracy}

\paragraph{Research questions} Now, we are interested in the metric accuracy, that is, agreement of AMR metrics with the human ratings. In particular, we would like to know, regarding:

\begin{itemize}
    \item Pair-wise parse accuracy: How do metrics agree with human preferences when ranking two candidates?
    \item Individual parse accuracy: Can metrics tell apart acceptable from unacceptable parses?
\end{itemize}

Note that these are hard tasks for metrics, since both T5 and \bart show performance levels on par or above estimated measurements for human IAA. Therefore, smaller structural divergences from the reference can potentially have a bigger impact on parse acceptability (or preference) than larger structural deviations, that could express different (but valid) interpretations or (near-)paraphrases. 

\subsection{Evaluation metrics}

\paragraph{Pairwise accuracy} Recall that the human assigned one of three ratings: 1, if AMR $x$ is better, $-1$, if AMR $y$ is better, and 0 if there is no considerable quality difference between two candidate graphs $x$ and $y$. A metric assigns two real values, $m(x, g)$ and $m(y, g)$, where $g$ is the reference graph. Mapping the score to $-1$ or $1$ is simple and intuitive, prompting us to introduce pair-wise accuracy. Consider a data set $SD$ that contains all graph triplets $(x, y, g)$ with a human preference sign (label $-1$ or $+1$). Further, let $\delta^m(x,y,g)=m(x,g) - m(y, g)$ the (signed) quality difference between $x$ and $y$ when using $m$. Analogously $\delta^h(x,y)$ is the human preference. Then, the pairwise accuracy is

\begin{equation}
    PA = \frac{1}{|D|}\sum_{(x,y,g) \in D} \mathbb{I}[\delta^m(x,y,g)\cdot \delta^h(x,y) > 0]
\end{equation}

This measures the ratio of candidate pairs where the metric has made the same signed decision as the human, in preferring one over the other parse. 

\paragraph{Acceptability score} When rating acceptability, the human rates a single parse (given its sentence), assigning 1 (acceptable) or 0 (no acceptable). The metrics make use of the reference graph to compute a score. Aiming at an evaluation metric that makes as few assumptions as possible, 
we formulate the following expectation for an AMR graph metric to fulfill: the average rank of the scores for parses that have been labeled acceptable by the human should surpass the average rank of the scores for parses labeled as being not acceptable. Let $\mathcal{I}^+$ ($\mathcal{I}^-$) be the set of indices for which the human has assigned a label that indicates (un-)acceptability. Let $S=\{m(X_1, G_1) ...m(X_n, G_n)\}$ be the metric $m$'s scores over all $(x, g)$ parse/reference pairs, and $R$ be the ranks of $D$. Let $R^+$ (and $R^-$) be the set of ranks indexed by $\mathcal{I}^+$ (and $\mathcal{I}^-$). Then

\begin{equation}
    \mathbb{A}\Delta = avg(R^+)-avg(R^-)
\end{equation}

To increase robustness, we use $avg$ := median.

\subsection{Results}

\begin{table}
    \centering
    \scalebox{0.8}{
    \begin{tabular}{lrrrr}
    & \multicolumn{2}{c}{Little Prince} & \multicolumn{2}{c}{AMR3} \\
         & PA & $\mathbb{A}\Delta$ & PA & $\mathbb{A}\Delta$ \\
         \toprule
         HUM   &   1.0 & 233& 1.0 & 234\\
         \midrule
         RAND   &   0.5 & 0.0& 0.5 & 0.0\\
         \midrule
        \simple & 0.66$^\dagger$  & 11.0 & 0.68$^\dagger$  & 39.5$^\dagger$ \\
        \sema & 0.66$^\dagger$ & 24.3& 0.7$^\dagger$  & 35.3$^\dagger$\\
        \sembleu-k2 & 0.67$^\dagger$  & 25.0 & 0.74$^\dagger$ & 28.0  \\
        \sembleu-k3 & 0.63$^\dagger$  & 32.0  & 0.68$^\dagger$  & 29.0\\
        \smatch & \textbf{0.72}$^\dagger$  & 42.0$^\dagger$ & 0.7$^\dagger$  & 35.0$^\dagger$ \\ 
        \sxmatch & \textbf{0.72}$^\dagger$  & 35.3 & 0.7$^\dagger$  & 42.3$^\dagger$\\
        \wlk & 0.66$^\dagger$ & 28.0 & 0.68$^\dagger$ & 41.5$^\dagger$\\
        \wwlk-k2 & 0.63$^\dagger$  & 20.5 & 0.73$^\dagger$ & 51.0$^\dagger$ \\
        \wwlk-k3e2n & 0.66$^\dagger$ & \textbf{48.0}$^\dagger$ & \textbf{0.76}$^\dagger$ & \textbf{57.0}$^\dagger$ \\
        \bottomrule
    \end{tabular}}
    \caption{Metric agreement with human. $\dagger$: random baseline (RAND) not contained in 95\% confidence interval.}
    \label{tab:metric_accuracy}
\end{table}

The results are shown in Table \ref{tab:metric_accuracy}. We conclude:

\paragraph{\textit{All metrics are suitable for pairwise-ranking of parses from high-performance parsers}} All metrics significantly outperform the random baseline with regard to the pair-wise ranking accuracy (PA). For Little Prince, \smatch and \sxmatch yield the best performance, while for AMR3, WWLK-k3e2n has the best performance (closely followed by \sembleu-k2). Among different metrics, however, the differences are not large enough to confidently recommend one metric over the other.

\paragraph{\textit{Parse acceptability rating is hard}} When tasked to rate parse acceptability ($\mathbb{A}\Delta$), all metrics show issues. For Little Prince, only \smatch and \wwlk-k3e2n significantly outperform the chance baseline, while for AMR3 all metrics are significantly above chance level, except  \sembleu. Overall, however, the differences are not large enough to confidently recommend one metric over the other. On both corpora,  best results are achieved with \wwlk-k3e2n (Little Prince: 48.0, AMR3: 57.0).

\paragraph{Control experiment of metrics} We additionally parse a subset of 50 sentences with an older parser \cite{flanigan-etal-2014-discriminative} that scores more than 20 points lower \smatch, when compared with IAA as estimated in \citet{banarescu-etal-2013-abstract}. All metrics (with the exception of \simple for one pair) correctly figure out all rankings and acceptability (according to the human, BART and T5 are preferred in all cases, except two cases where all three systems deliver equally valid graphs). This indicates that metrics indeed can accurately tell apart quality differences, \textit{if} they are large enough and do not lie beyond human IAA. 

\section{Metric specificity}
\label{sec:metricspecificity}

We found little evidence that could help us giving recommendations on which metrics to prefer over others for monolingual parser evaluation in the high-performance regime. On the contrary, we found  evidence that no metric can sufficiently assess parse acceptability. Therefore, it is interesting to see whether the metrics can provide \textit{different} views on parse quality.

\subsection{Correlation analysis}

\paragraph{Statistics} We compute Spearman's $\rho$ over metric pairs. Spearman's $\rho$ calculates Pearson's $\rho$ on the ranked predictions, which increases robustness.

\begin{figure}
    \centering
    \includegraphics[width=0.8\linewidth]{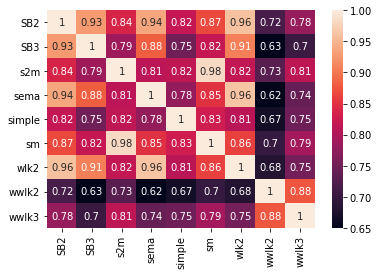}
    \caption{Inter-metric correlation on Little Prince.}
    \label{fig:lpcorr}
\end{figure}

\begin{figure}
    \centering
    \includegraphics[width=0.8\linewidth]{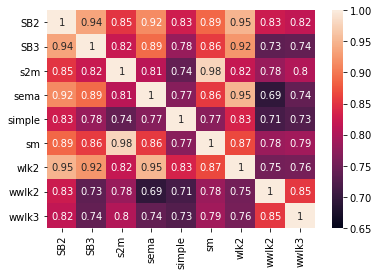}
    \caption{Inter-metric correlation on AMR3.}
    \label{fig:amr3corr}
\end{figure}

\paragraph{Results} Results are plotted in Figures \ref{fig:lpcorr} and \ref{fig:amr3corr}. For both datasets, we see that the Wasserstein metrics provide rankings that differ more from the rankings assigned by other metrics, suggesting that they have unique features. On the other hand, the \sembleu metrics tend to agree the most with the rankings of the other metrics, suggesting that they share more features with other metrics. On a pair-wise level, the most similar metrics are \smatch and \sxmatch, which is intuitive, since \sxmatch is an adaption of \smatch that also targets the comparison of AMRs from different sentences. Indeed, synonyms and similar concepts are unlikely to often occur in monolingual parsing, where parses contain exactly matching concepts. Further, \wlk very much agrees with \sembleu, which seems intuitive, since both aim at comparing larger AMR subgraphs. Lowest agreement is exhibited between \sema and \wwlk, perhaps because these metrics are of different complexity and share different goals: simple and fast match of structures vs.\ graded assessment for general AMR similarity. 

\section{Discussion of study limitations}
\label{sec:limitations}

There are limitations of our study:

\paragraph{Limitation I: \textit{Single vs.\ Double annotation}} While our quality annotations stem from an experienced human annotator, we would have liked to obtain annotations from a second annotator to measure IAA for AMR quality rating. This was partly precluded by the high costs of AMR annotation, which requires much time and experience. This is also reflected in the AMR benchmark corpora: the majority of graphs were created by a single annotator. Note, however, that some findings are independent of annotation (e.g., macro vs.\ micro metric corpus scoring, metric specificity).

\paragraph{Limitation II: \textit{Assessing individual suitability of metrics for rating high-performance parsers}} Our study reports relevant findings on (monolingual) AMR parsing evaluation in high-performance regimes, and on upper bounds of AMR parsing. But an important question we had to leave open is the individual suitability of the metrics for comparing high-performance parsers.
    
\paragraph{Limitation III: \textit{Single-reference parses and ambiguity}} Elaborating on \textit{Limitation II} and recalling that AMR benchmarks have only single references, another caveat is that potentially correct metric behavior may be misinterpreted in our study. E.g., if a sentence allows two different interpretations, a metric might (correctly) yield a low score for the reference (different meaning), while the (reference-less) human rating may find the parse acceptable. This issue may also be mitigated by providing (costly) double annotation of AMR benchmark sentences.

\vspace*{2mm}
To facilitate follow-up research, we release the annotated data. Our Little Prince annotations can be freely released, while AMR3 annotations require proof of LDC license.

\section{Discussion and Conclusions}
\label{sec:recommendations}

\paragraph{Main recommendations} based on our study:

\begin{description}
    \item[\textit{Recommendation I}] Besides micro aggregate scores we recommend using \textbf{a macro aggregate score} for parse evaluation (e.g., macro \smatch, computed as an average over sentence scores): Commonly, only micro corpus statistics are used to compare and rank parsers. Yet, we found that macro (sentence-average) metrics can provide a valuable \textbf{complementary assessment} that can highlight important \textit{additional} strengths of high-performance parsers. \vspace*{-1.5ex}
    \item[\textit{Recommendation II}] We recommend conducting \textbf{more human evaluation of AMR parses}. With the available high-performance AMR parsers, it becomes more important to conduct manual analyses of parse quality. Our study provides evidence that AMR parsing still has large room for improvement, due to small but significant errors. Since this may not be noticeable for (current) metrics when given a single human reference, \textbf{future work on parsing may profit from careful human acceptability assessments}.
\end{description}

\paragraph{T5 vs.\ \bart: which parser to prefer?} Next to  AMR parser developers, this question mainly concerns potential users of AMR parsers. Fine-tuned T5 and \bart are both powerful AMR parsers. We observe a slight tendency that researchers prefer \bart, possibly since it achieves slightly better \smatch scores than T5 on the AMR3 benchmark. But our work shows that differences between the systems are often finer than what can be assessed with structural overlap metrics (\smatch), and both systems are generally strong but struggle with small but significant meaning errors. 

In our study we found that when choosing between T5 and \bart based AMR systems, \textbf{the choice might depend on the target domain}. Indeed, our results on Little Prince and AMR3 (mainly news) could indicate that \textbf{T5 may have an edge over \bart when parsing literary texts}, and shorter sentences in general, while \textbf{\bart has an edge over T5 when parsing longer sentences, and sentences from news sources}, especially if they are longer. However, it must be clearly noted, that we do not know (yet) whether this insight carries over to other types of literary texts. 

Perhaps, if we presume that performance is carried over to other types of literary texts, a possible explanation can be found in the data these two large models were trained on. \bart uses the same training data as RoBERTa \cite{liu2019roberta}, e.g., Wikipedia, book corpora and news. T5 leverages the colossal common crawl corpus (C4), that contains all kinds of texts scraped from the web. This \textit{could} make T5 more robust to AMR domain changes, but less suitable for analysing longer sentences, since these may occur more frequently in \bart's corpora that seem more normative.

\paragraph{Which AMR metric to use?} \textbf{Our findings do not provide conclusive evidence} on this question, partly due to insufficient data size, partly due to the general difficulty of the task. \wwlk-k3e2n seems slightly more useful for detecting parse acceptability and pairwise ranking on news, while \smatch yields best ranking on Little Prince. 

However, our work shows that \textbf{it can be useful to calculate more than one metric to compare parsers}. In particular, we saw that predictions of structural matching metrics differ considerably from graded semantic similarity-based metrics, such as the WWLK metric variants. This suggests that these two types can provide complementary perspectives on parsing accuracy. Metric selection may, of course, also be driven by users' specific desiderata, such as speed (\sema, \sembleu, \wlk), 1-1 alignment  (\smatch), n:m alignment (\wwlk), or graded matching (\smatch, \wwlk). Overall, we see much \textbf{profit to gain from more research into AMR metrics}, and will now outline a direction that we believe is very interesting.

\paragraph{A direction for future research: Reference-less AMR metrics}
   
Recall that for human quality assessments a candidate graph is compared to a \textit{sentence}, in lieu of a reference AMR. If this process can be approximated by a metric, we gain an important mechanism for assessing the quality of high-performance parsers: a measure that is cheap and not biased towards a single reference.

To date, referenceless AMR parse quality rating has been attempted by \citet{opitz-frank-2019-automatic, opitz-2020-amr}. However, an unsolved issue is that this approach does not approximate a human quality assessment, but instead tries to project \smatch score without using a reference, and we saw that \smatch cannot well assess the impact of fine errors of high-performing parsers. 

A worthwhile solution could be found in the exploitation of indirect tools: E.g., our human annotation indicated that significant, but small structural errors are sometimes due to coreference, which is known to be a hard task in general \cite{levesque2012winograd} and for AMR in particular \cite{anikina-etal-2020-predicting}. Therefore, e.g., one may profit from matching parses from a high-performance parser against the structures predicted by a strong coreference system, possibly with the help of a predicted AMR-to-text alignment \cite{blodgett-schneider-2021-probabilistic}. Another promising route to take may be to invert approaches of \citet{opitz2020towards, manning-schneider-2021-referenceless} who evaluate AMR-to-text generation without reliance on a reference by using a strong parser for back-parsing. It may be beneficial to use strong AMR-to-text systems to generate from candidate AMRs, and to match the generations against the source sentence using strong automatic text-to-text metrics.

\section*{Acknowledgements} We are grateful to three anonymous reviewers for their valuable comments that have helped to improve this paper. We are also thankful to Julius Steen for valuable discussions.

\bibliography{anthology,custom}

\begin{thebibliography}{41}
\expandafter\ifx\csname natexlab\endcsname\relax\def\natexlab#1{#1}\fi

\bibitem[{Anchi\^{e}ta et~al.(2019)Anchi\^{e}ta, Cabezudo, and
  Pardo}]{anchieta2019sema}
Rafael~Torres Anchi\^{e}ta, Marco Antonio~Sobrevilla Cabezudo, and Thiago
  Alexandre~Salgueiro Pardo. 2019.
\newblock Sema: an extended semantic evaluation for amr.
\newblock In \emph{(To appear) Proceedings of the 20th Computational
  Linguistics and Intelligent Text Processing}. Springer International
  Publishg.

\bibitem[{Anikina et~al.(2020)Anikina, Koller, and
  Roth}]{anikina-etal-2020-predicting}
Tatiana Anikina, Alexander Koller, and Michael Roth. 2020.
\newblock \href {https://aclanthology.org/2020.crac-1.4} {Predicting
  coreference in {A}bstract {M}eaning {R}epresentations}.
\newblock In \emph{Proceedings of the Third Workshop on Computational Models of
  Reference, Anaphora and Coreference}, pages 33--38, Barcelona, Spain
  (online). Association for Computational Linguistics.

\bibitem[{Banarescu et~al.(2013)Banarescu, Bonial, Cai, Georgescu, Griffitt,
  Hermjakob, Knight, Koehn, Palmer, and
  Schneider}]{banarescu-etal-2013-abstract}
Laura Banarescu, Claire Bonial, Shu Cai, Madalina Georgescu, Kira Griffitt, Ulf
  Hermjakob, Kevin Knight, Philipp Koehn, Martha Palmer, and Nathan Schneider.
  2013.
\newblock \href {https://www.aclweb.org/anthology/W13-2322} {Abstract meaning
  representation for sembanking}.
\newblock In \emph{Proceedings of the 7th Linguistic Annotation Workshop and
  Interoperability with Discourse}, pages 178--186, Sofia, Bulgaria.
  Association for Computational Linguistics.

\bibitem[{Beck et~al.(2018)Beck, Haffari, and Cohn}]{beck-etal-2018-graph}
Daniel Beck, Gholamreza Haffari, and Trevor Cohn. 2018.
\newblock \href {https://doi.org/10.18653/v1/P18-1026} {Graph-to-sequence
  learning using gated graph neural networks}.
\newblock In \emph{Proceedings of the 56th Annual Meeting of the Association
  for Computational Linguistics (Volume 1: Long Papers)}, pages 273--283,
  Melbourne, Australia. Association for Computational Linguistics.

\bibitem[{Bevilacqua et~al.(2021)Bevilacqua, Blloshmi, and
  Navigli}]{bevilacqua2021one}
Michele Bevilacqua, Rexhina Blloshmi, and Roberto Navigli. 2021.
\newblock One spring to rule them both: Symmetric amr semantic parsing and
  generation without a complex pipeline.
\newblock In \emph{Proceedings of the AAAI Conference on Artificial
  Intelligence}, volume~35, pages 12564--12573.

\bibitem[{Blodgett and Schneider(2021)}]{blodgett-schneider-2021-probabilistic}
Austin Blodgett and Nathan Schneider. 2021.
\newblock \href {https://doi.org/10.18653/v1/2021.acl-long.257} {Probabilistic,
  structure-aware algorithms for improved variety, accuracy, and coverage of
  {AMR} alignments}.
\newblock In \emph{Proceedings of the 59th Annual Meeting of the Association
  for Computational Linguistics and the 11th International Joint Conference on
  Natural Language Processing (Volume 1: Long Papers)}, pages 3310--3321,
  Online. Association for Computational Linguistics.

\bibitem[{Cai and Lam(2020)}]{cai-lam-2020-amr}
Deng Cai and Wai Lam. 2020.
\newblock \href {https://doi.org/10.18653/v1/2020.acl-main.119} {{AMR} parsing
  via graph-sequence iterative inference}.
\newblock In \emph{Proceedings of the 58th Annual Meeting of the Association
  for Computational Linguistics}, pages 1290--1301, Online. Association for
  Computational Linguistics.

\bibitem[{Cai and Knight(2013)}]{cai-knight-2013-smatch}
Shu Cai and Kevin Knight. 2013.
\newblock \href {https://www.aclweb.org/anthology/P13-2131} {{S}match: an
  evaluation metric for semantic feature structures}.
\newblock In \emph{Proceedings of the 51st Annual Meeting of the Association
  for Computational Linguistics (Volume 2: Short Papers)}, pages 748--752,
  Sofia, Bulgaria. Association for Computational Linguistics.

\bibitem[{Dohare et~al.(2017)Dohare, Karnick, and Gupta}]{dohare2017text}
Shibhansh Dohare, Harish Karnick, and Vivek Gupta. 2017.
\newblock \href {http://arxiv.org/abs/1706.01678} {Text summarization using
  abstract meaning representation}.

\bibitem[{Flanigan et~al.(2014)Flanigan, Thomson, Carbonell, Dyer, and
  Smith}]{flanigan-etal-2014-discriminative}
Jeffrey Flanigan, Sam Thomson, Jaime Carbonell, Chris Dyer, and Noah~A. Smith.
  2014.
\newblock \href {https://doi.org/10.3115/v1/P14-1134} {A discriminative
  graph-based parser for the {A}bstract {M}eaning {R}epresentation}.
\newblock In \emph{Proceedings of the 52nd Annual Meeting of the Association
  for Computational Linguistics (Volume 1: Long Papers)}, pages 1426--1436,
  Baltimore, Maryland. Association for Computational Linguistics.

\bibitem[{Freitag et~al.(2021)Freitag, Rei, Mathur, Lo, Stewart, Foster, Lavie,
  and Bojar}]{freitag-etal-2021-results}
Markus Freitag, Ricardo Rei, Nitika Mathur, Chi-kiu Lo, Craig Stewart, George
  Foster, Alon Lavie, and Ond{\v{r}}ej Bojar. 2021.
\newblock \href {https://aclanthology.org/2021.wmt-1.73} {Results of the
  {WMT}21 metrics shared task: Evaluating metrics with expert-based human
  evaluations on {TED} and news domain}.
\newblock In \emph{Proceedings of the Sixth Conference on Machine Translation},
  pages 733--774, Online. Association for Computational Linguistics.

\bibitem[{Levesque et~al.(2012)Levesque, Davis, and
  Morgenstern}]{levesque2012winograd}
Hector Levesque, Ernest Davis, and Leora Morgenstern. 2012.
\newblock The winograd schema challenge.
\newblock In \emph{Thirteenth international conference on the principles of
  knowledge representation and reasoning}.

\bibitem[{Levi(1942)}]{levi1942finite}
Friedrich~Wilhelm Levi. 1942.
\newblock \emph{Finite geometrical systems: six public lectues delivered in
  February, 1940, at the University of Calcutta}.
\newblock University of Calcutta.

\bibitem[{Lewis et~al.(2020)Lewis, Liu, Goyal, Ghazvininejad, Mohamed, Levy,
  Stoyanov, and Zettlemoyer}]{lewis-etal-2020-bart}
Mike Lewis, Yinhan Liu, Naman Goyal, Marjan Ghazvininejad, Abdelrahman Mohamed,
  Omer Levy, Veselin Stoyanov, and Luke Zettlemoyer. 2020.
\newblock \href {https://doi.org/10.18653/v1/2020.acl-main.703} {{BART}:
  Denoising sequence-to-sequence pre-training for natural language generation,
  translation, and comprehension}.
\newblock In \emph{Proceedings of the 58th Annual Meeting of the Association
  for Computational Linguistics}, pages 7871--7880, Online. Association for
  Computational Linguistics.

\bibitem[{Liao et~al.(2018)Liao, Lebanoff, and Liu}]{liao-etal-2018-abstract}
Kexin Liao, Logan Lebanoff, and Fei Liu. 2018.
\newblock \href {https://aclanthology.org/C18-1101} {{A}bstract {M}eaning
  {R}epresentation for multi-document summarization}.
\newblock In \emph{Proceedings of the 27th International Conference on
  Computational Linguistics}, pages 1178--1190, Santa Fe, New Mexico, USA.
  Association for Computational Linguistics.

\bibitem[{Liu et~al.(2019)Liu, Ott, Goyal, Du, Joshi, Chen, Levy, Lewis,
  Zettlemoyer, and Stoyanov}]{liu2019roberta}
Yinhan Liu, Myle Ott, Naman Goyal, Jingfei Du, Mandar Joshi, Danqi Chen, Omer
  Levy, Mike Lewis, Luke Zettlemoyer, and Veselin Stoyanov. 2019.
\newblock Roberta: A robustly optimized bert pretraining approach.
\newblock \emph{arXiv preprint arXiv:1907.11692}.

\bibitem[{Lyu and Titov(2018)}]{lyu-titov-2018-amr}
Chunchuan Lyu and Ivan Titov. 2018.
\newblock \href {https://doi.org/10.18653/v1/P18-1037} {{AMR} parsing as graph
  prediction with latent alignment}.
\newblock In \emph{Proceedings of the 56th Annual Meeting of the Association
  for Computational Linguistics (Volume 1: Long Papers)}, pages 397--407,
  Melbourne, Australia. Association for Computational Linguistics.

\bibitem[{Ma et~al.(2019)Ma, Wei, Bojar, and Graham}]{ma-etal-2019-results}
Qingsong Ma, Johnny Wei, Ond{\v{r}}ej Bojar, and Yvette Graham. 2019.
\newblock \href {https://doi.org/10.18653/v1/W19-5302} {Results of the {WMT}19
  metrics shared task: Segment-level and strong {MT} systems pose big
  challenges}.
\newblock In \emph{Proceedings of the Fourth Conference on Machine Translation
  (Volume 2: Shared Task Papers, Day 1)}, pages 62--90, Florence, Italy.
  Association for Computational Linguistics.

\bibitem[{Manning and Schneider(2021)}]{manning-schneider-2021-referenceless}
Emma Manning and Nathan Schneider. 2021.
\newblock \href {https://aclanthology.org/2021.eval4nlp-1.12} {Referenceless
  parsing-based evaluation of {AMR}-to-{E}nglish generation}.
\newblock In \emph{Proceedings of the 2nd Workshop on Evaluation and Comparison
  of NLP Systems}, pages 114--122, Punta Cana, Dominican Republic. Association
  for Computational Linguistics.

\bibitem[{Mathur et~al.(2020)Mathur, Wei, Freitag, Ma, and
  Bojar}]{mathur-etal-2020-results}
Nitika Mathur, Johnny Wei, Markus Freitag, Qingsong Ma, and Ond{\v{r}}ej Bojar.
  2020.
\newblock \href {https://aclanthology.org/2020.wmt-1.77} {Results of the
  {WMT}20 metrics shared task}.
\newblock In \emph{Proceedings of the Fifth Conference on Machine Translation},
  pages 688--725, Online. Association for Computational Linguistics.

\bibitem[{Opitz(2020)}]{opitz-2020-amr}
Juri Opitz. 2020.
\newblock \href {https://aclanthology.org/2020.aacl-main.27} {{AMR} quality
  rating with a lightweight {CNN}}.
\newblock In \emph{Proceedings of the 1st Conference of the Asia-Pacific
  Chapter of the Association for Computational Linguistics and the 10th
  International Joint Conference on Natural Language Processing}, pages
  235--247, Suzhou, China. Association for Computational Linguistics.

\bibitem[{Opitz et~al.(2021)Opitz, Daza, and Frank}]{opitz2021weisfeiler}
Juri Opitz, Angel Daza, and Anette Frank. 2021.
\newblock \href {https://doi.org/10.1162/tacl_a_00435} {{Weisfeiler-Leman in
  the Bamboo: Novel AMR Graph Metrics and a Benchmark for AMR Graph
  Similarity}}.
\newblock \emph{Transactions of the Association for Computational Linguistics},
  9:1425--1441.

\bibitem[{Opitz and Frank(2019)}]{opitz-frank-2019-automatic}
Juri Opitz and Anette Frank. 2019.
\newblock \href {https://doi.org/10.18653/v1/S19-1024} {Automatic accuracy
  prediction for {AMR} parsing}.
\newblock In \emph{Proceedings of the Eighth Joint Conference on Lexical and
  Computational Semantics (*{SEM} 2019)}, pages 212--223, Minneapolis,
  Minnesota. Association for Computational Linguistics.

\bibitem[{Opitz and Frank(2021)}]{opitz2020towards}
Juri Opitz and Anette Frank. 2021.
\newblock \href {https://www.aclweb.org/anthology/2021.eacl-main.129} {Towards
  a decomposable metric for explainable evaluation of text generation from
  {AMR}}.
\newblock In \emph{Proceedings of the 16th Conference of the European Chapter
  of the Association for Computational Linguistics: Main Volume}, pages
  1504--1518, Online. Association for Computational Linguistics.

\bibitem[{Opitz and Frank(2022)}]{opitz2022sbert}
Juri Opitz and Anette Frank. 2022.
\newblock Sbert studies meaning representations: Decomposing sentence
  embeddings into explainable amr meaning features.
\newblock \emph{arXiv preprint arXiv:2206.07023}.

\bibitem[{Opitz et~al.(2020)Opitz, Parcalabescu, and Frank}]{opitz-tacl}
Juri Opitz, Letitia Parcalabescu, and Anette Frank. 2020.
\newblock \href {https://doi.org/10.1162/tacl\_a\_00329} {Amr similarity
  metrics from principles}.
\newblock \emph{Transactions of the Association for Computational Linguistics},
  8:522--538.

\bibitem[{Palmer et~al.(2005)Palmer, Gildea, and
  Kingsbury}]{palmer-etal-2005-proposition}
Martha Palmer, Daniel Gildea, and Paul Kingsbury. 2005.
\newblock \href {https://doi.org/10.1162/0891201053630264} {The proposition
  bank: An annotated corpus of semantic roles}.
\newblock \emph{Computational Linguistics}, 31(1):71--106.

\bibitem[{Papineni et~al.(2002)Papineni, Roukos, Ward, and
  Zhu}]{papineni-etal-2002-bleu}
Kishore Papineni, Salim Roukos, Todd Ward, and Wei-Jing Zhu. 2002.
\newblock \href {https://doi.org/10.3115/1073083.1073135} {{B}leu: a method for
  automatic evaluation of machine translation}.
\newblock In \emph{Proceedings of the 40th Annual Meeting of the Association
  for Computational Linguistics}, pages 311--318, Philadelphia, Pennsylvania,
  USA. Association for Computational Linguistics.

\bibitem[{Pennington et~al.(2014)Pennington, Socher, and
  Manning}]{pennington-etal-2014-glove}
Jeffrey Pennington, Richard Socher, and Christopher Manning. 2014.
\newblock \href {https://doi.org/10.3115/v1/D14-1162} {{G}love: Global vectors
  for word representation}.
\newblock In \emph{Proceedings of the 2014 Conference on Empirical Methods in
  Natural Language Processing ({EMNLP})}, pages 1532--1543, Doha, Qatar.
  Association for Computational Linguistics.

\bibitem[{Raffel et~al.(2019)Raffel, Shazeer, Roberts, Lee, Narang, Matena,
  Zhou, Li, and Liu}]{T5}
Colin Raffel, Noam Shazeer, Adam Roberts, Katherine Lee, Sharan Narang, Michael
  Matena, Yanqi Zhou, Wei Li, and Peter~J. Liu. 2019.
\newblock \href {http://arxiv.org/abs/1910.10683} {Exploring the limits of
  transfer learning with a unified text-to-text transformer}.
\newblock \emph{CoRR}, abs/1910.10683.

\bibitem[{Ribeiro et~al.(2022)Ribeiro, Liu, Gurevych, Dreyer, and
  Bansal}]{ribeiro2022factgraph}
Leonardo Ribeiro, Mengwen Liu, Iryna Gurevych, Markus Dreyer, and Mohit Bansal.
  2022.
\newblock \href {https://aclanthology.org/2022.naacl-main.236} {{F}act{G}raph:
  Evaluating factuality in summarization with semantic graph representations}.
\newblock In \emph{Proceedings of the 2022 Conference of the North American
  Chapter of the Association for Computational Linguistics: Human Language
  Technologies}, pages 3238--3253, Seattle, United States. Association for
  Computational Linguistics.

\bibitem[{Ribeiro et~al.(2019)Ribeiro, Gardent, and
  Gurevych}]{ribeiro-etal-2019-enhancing}
Leonardo F.~R. Ribeiro, Claire Gardent, and Iryna Gurevych. 2019.
\newblock \href {https://doi.org/10.18653/v1/D19-1314} {Enhancing {AMR}-to-text
  generation with dual graph representations}.
\newblock In \emph{Proceedings of the 2019 Conference on Empirical Methods in
  Natural Language Processing and the 9th International Joint Conference on
  Natural Language Processing (EMNLP-IJCNLP)}, pages 3183--3194, Hong Kong,
  China. Association for Computational Linguistics.

\bibitem[{Sellam et~al.(2020)Sellam, Das, and Parikh}]{sellam-etal-2020-bleurt}
Thibault Sellam, Dipanjan Das, and Ankur Parikh. 2020.
\newblock \href {https://doi.org/10.18653/v1/2020.acl-main.704} {{BLEURT}:
  Learning robust metrics for text generation}.
\newblock In \emph{Proceedings of the 58th Annual Meeting of the Association
  for Computational Linguistics}, pages 7881--7892, Online. Association for
  Computational Linguistics.

\bibitem[{Shervashidze et~al.(2011)Shervashidze, Schweitzer, Van~Leeuwen,
  Mehlhorn, and Borgwardt}]{shervashidze2011weisfeiler}
Nino Shervashidze, Pascal Schweitzer, Erik~Jan Van~Leeuwen, Kurt Mehlhorn, and
  Karsten~M Borgwardt. 2011.
\newblock Weisfeiler-lehman graph kernels.
\newblock \emph{Journal of Machine Learning Research}, 12(9).

\bibitem[{Song and Gildea(2019)}]{song-gildea-2019-sembleu}
Linfeng Song and Daniel Gildea. 2019.
\newblock \href {https://doi.org/10.18653/v1/P19-1446} {{S}em{B}leu: A robust
  metric for {AMR} parsing evaluation}.
\newblock In \emph{Proceedings of the 57th Annual Meeting of the Association
  for Computational Linguistics}, pages 4547--4552, Florence, Italy.
  Association for Computational Linguistics.

\bibitem[{Song et~al.(2019)Song, Gildea, Zhang, Wang, and
  Su}]{10.1162/tacl_a_00252}
Linfeng Song, Daniel Gildea, Yue Zhang, Zhiguo Wang, and Jinsong Su. 2019.
\newblock \href {https://doi.org/10.1162/tacl_a_00252} {{Semantic Neural
  Machine Translation Using AMR}}.
\newblock \emph{Transactions of the Association for Computational Linguistics},
  7:19--31.

\bibitem[{Togninalli et~al.(2019)Togninalli, Ghisu, Llinares-L\'{o}pez, Rieck,
  and Borgwardt}]{NEURIPS2019_73fed7fd}
Matteo Togninalli, Elisabetta Ghisu, Felipe Llinares-L\'{o}pez, Bastian Rieck,
  and Karsten Borgwardt. 2019.
\newblock \href
  {https://proceedings.neurips.cc/paper/2019/file/73fed7fd472e502d8908794430511f4d-Paper.pdf}
  {Wasserstein weisfeiler-lehman graph kernels}.
\newblock In \emph{Advances in Neural Information Processing Systems},
  volume~32, pages 6436--6446. Curran Associates, Inc.

\bibitem[{Uhrig et~al.(2021)Uhrig, Garcia, Opitz, and
  Frank}]{uhrig-etal-2021-translate}
Sarah Uhrig, Yoalli Garcia, Juri Opitz, and Anette Frank. 2021.
\newblock \href {https://doi.org/10.18653/v1/2021.iwpt-1.6} {Translate, then
  parse! a strong baseline for cross-lingual {AMR} parsing}.
\newblock In \emph{Proceedings of the 17th International Conference on Parsing
  Technologies and the IWPT 2021 Shared Task on Parsing into Enhanced Universal
  Dependencies (IWPT 2021)}, pages 58--64, Online. Association for
  Computational Linguistics.

\bibitem[{Wang et~al.(2015)Wang, Xue, and Pradhan}]{wang-etal-2015-transition}
Chuan Wang, Nianwen Xue, and Sameer Pradhan. 2015.
\newblock \href {https://doi.org/10.3115/v1/N15-1040} {A transition-based
  algorithm for {AMR} parsing}.
\newblock In \emph{Proceedings of the 2015 Conference of the North {A}merican
  Chapter of the Association for Computational Linguistics: Human Language
  Technologies}, pages 366--375, Denver, Colorado. Association for
  Computational Linguistics.

\bibitem[{Xu et~al.(2020)Xu, Li, Zhu, Zhang, and Zhou}]{xu-etal-2020-improving}
Dongqin Xu, Junhui Li, Muhua Zhu, Min Zhang, and Guodong Zhou. 2020.
\newblock \href {https://doi.org/10.18653/v1/2020.emnlp-main.196} {Improving
  {AMR} parsing with sequence-to-sequence pre-training}.
\newblock In \emph{Proceedings of the 2020 Conference on Empirical Methods in
  Natural Language Processing (EMNLP)}, pages 2501--2511, Online. Association
  for Computational Linguistics.

\bibitem[{Zeidler et~al.(2022)Zeidler, Opitz, and
  Frank}]{zeidler-etal-2022-dynamic}
Laura Zeidler, Juri Opitz, and Anette Frank. 2022.
\newblock \href {https://doi.org/10.18653/v1/2022.starsem-1.14} {A dynamic,
  interpreted {C}heck{L}ist for meaning-oriented {NLG} metric evaluation {--}
  through the lens of semantic similarity rating}.
\newblock In \emph{Proceedings of the 11th Joint Conference on Lexical and
  Computational Semantics}, pages 157--172, Seattle, Washington. Association
  for Computational Linguistics.

\end{thebibliography}
\bibliographystyle{acl_natbib}

\appendix

\section{Appendix}
\label{app:appendix}

\subsection{Sentence length vs.\ score}
\label{app:sentlen}

\begin{figure}[h]
    \centering
    \includegraphics[width=\linewidth]{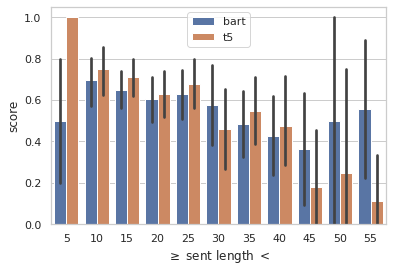}
    \caption{\textbf{Sentence length vs.\ human acceptability} on all annotated data. 55 includes all sentences longer than 55 tokens. See Figure \ref{fig:sentlencounts} for occurences of different sentence lengths.}
    \label{fig:slenhum}
\end{figure}

\begin{figure}[h]
    \centering
    \includegraphics[width=\linewidth]{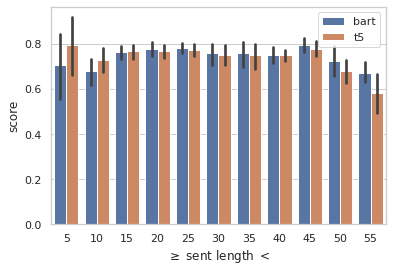}
    \caption{\textbf{Sentence length vs.\ Smatch} on all annotated data. 55 includes all sentences longer than 55 tokens. See Figure \ref{fig:sentlencounts} for occurences of different sentence lengths. Other metrics look similar.}
    \label{fig:slensm}
\end{figure}

\begin{figure}[h]
    \centering
    \includegraphics[width=\linewidth]{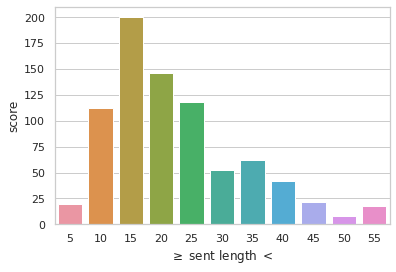}
    \caption{Sentence length occurrences. 55 includes all sentences longer than 55 tokens.}
    \label{fig:sentlencounts}
\end{figure}

See Figures \ref{fig:slenhum}, \ref{fig:slensm}. For total sentence length distribution see Figure \ref{fig:sentlencounts}.

%This is an appendix.

\end{document}